\documentclass[conference]{IEEEtran}
\IEEEoverridecommandlockouts
\usepackage{cite}
\usepackage{amsmath,amssymb,amsfonts}
\usepackage{algorithmic}
\usepackage{graphicx}
\usepackage{textcomp}
\usepackage{xcolor}
\usepackage{multirow}
\usepackage{algorithm}
\usepackage{algorithmic}
\usepackage{amsmath}
\usepackage{breqn}

\def\BibTeX{{\rm B\kern-.05em{\sc i\kern-.025em b}\kern-.08em
    T\kern-.1667em\lower.7ex\hbox{E}\kern-.125emX}}
\begin{document}

\title{Towards CGAN-based Satellite Image Synthesis with Partial Pixel-Wise Annotation}

\author{\IEEEauthorblockN{Hadi Mansourifar}
\IEEEauthorblockA{\small Systems Engineering Department \\ Colorado State University \\ Fort Collins \\
hadiman@colostate.edu}
\and
\IEEEauthorblockN{Steven J. Simske}
\IEEEauthorblockA{\small Systems Engineering Department \\ Colorado State University \\ Fort Collins \\
steve.simske@colostate.edu}
\and

}

\maketitle

\begin{abstract}
Conditional Generative Adversarial Nets (CGANs) need a significantly huge dataset with a detailed pixel-wise annotation to generate high-quality images. Unfortunately, any amount of missing pixel annotations may significantly impact the result not only locally, but also in annotated areas. To the best of our knowledge, such a challenge has never been investigated in the broader field of GANs. In this paper, we take the first step in this direction to study the problem of CGAN-based satellite image synthesis given partially annotated images. We first define the problem of image synthesis using partially annotated data, and we discuss a scenario in which we face such a challenge. We then propose an effective solution called detail augmentation to address this problem. To do so, we tested two different approaches to augment details to compensate for missing pixel-wise annotations. In the first approach, we augmented the original images with their Canny edges to using the CGAN to compensate for the missing annotations. The second approach, however, attempted to assign a color to all pixels with missing annotation. Eventually, a different CGAN was trained to translate the new feature images into a final output. 
\end{abstract}

\begin{IEEEkeywords}
CGANs, Satellite Imaging, Partially Annotated Data
\end{IEEEkeywords}

\section{Introduction}
Adversarial image synthesis \cite{mansourifar2022gan,isola2017image,zhu2017unpaired,zhang2020enhanced,mansourifar2019virtual} is one of the hot topics in AI, with many successful use cases being reported. However, the Partial Pixel-wise Annotation (PPA) problem has never been investigated in the field of Conditional Generative Adversarial Nets (CGANs). Studying the PPA in the context of employing CGANs is of great importance because of the following reasons: (i) collecting Full Pixel-wise Annotation (FPA) is extremely costly and cumbersome, especially in high-resolution images like remote sensing data; (ii) even in the case of an available FPA for training the CGAN, it is inconvenient for the end users to provide an FPA during testing via an interactive user interface; and (iii) all of the pixel-wise datasets in the satellite imaging domain lack FPA and have not originally been collected for the aim of image synthesis. Unfortunately, ordinary CGAN training with PPA needs to meet quality expectations in terms of object detection scores. This deficiency is due to the following reasons:
\begin{itemize}
\item Imbalanced nature of PFA: the objects of interest in available satellite image datasets are limited \cite{waqas2019isaid}. In other words,  pixels with no annotation significantly outnumber annotated pixels.
\item The generalization - power trade-off is more conspicuous in the face of the PPA problem. It means that more powerful GAN architectures in normal cases are more vulnerable in the face of PPA scenarios.  
\end{itemize}

In this paper, we take a first step to address these aforementioned challenges. To do so, we propose a new approach called CGAN-based detail augmentation. The proposed detail augmentation is implemented with two variations: (i) Partial Detail Augmentation (PDA) and (ii) Full Detail Augmentation (FDA). In PDA, a pix2pix model is trained to augment the original input feature images with their Canny edges. In FDA, however, a different pix2pix model is trained to augment the segmented regions to the original input feature image. Finally, the obtained feature images are translated to the output using a second pix2pix model \cite{isola2017image}. Our experimental results show that the proposed method can significantly improve the object detection scores using several well-known object detector models, including AWS Rekognition, Google Vision Cloud, and YOLOv3. To do so, we select a set of labels like aircraft, buildings, etc. Afterward, we compare the impact of the proposed models using the confidence scores corresponding to each label with or without detail augmentation. Our contributions are as follows.
\begin{itemize}
\item For the first time in Machine Learning and  Computer Vision, we investigate the Partial Pixel-wise Annotation (PPA) problem in GAN-based image synthesis.

\item We demonstrate the scenarios in which the PPA problem might be encountered.

\item We propose two different approaches to tackle the PPA in the CGANs called Partial Detail Augmentation (PDA) and Full Detail Augmentation (FDA).

\item We demonstrate that PDA and FDA can improve object detection scores using independent object detection models.
\item We demonstrate that FDA can outperform PDA in terms of weak object detection scores.
\end{itemize}

\begin{table*}
\centering
\caption{A summary of CGAN-based satellite image synthesis related works.}
\label{map}
\begin{tabular}{|c|c|c|c|c|l|} 
\hline
\textbf{Type}                & \textbf{Author}           & \textbf{Dataset} & \textbf{Year} & \textbf{GAN Type} & \multicolumn{1}{c|}{\textbf{Comment}}                                                                                                                                                                                                                                                                                                                           \\ 
\hline
Map Synthesis                & \textit{Isola et al. \cite{isola2017image}}     & Google Map       & 2017          & CGAN              & \begin{tabular}[c]{@{}l@{}}\textit{Used pix2pix for the first time for map synthesis given paired}\\\textit{training data}\end{tabular}                                                                                                                                                                                                                         \\ 
\hline
Map Synthesis                & \textit{Zhu et al. \cite{zhu2017unpaired}}       & Google Map       & 2017          & CycleGAN          & \begin{tabular}[c]{@{}l@{}}\textit{Used CycleGAN to translate between domains with unpaired}\\\textit{input-output examples}\end{tabular}                                                                                                                                                                                                                      \\ 
\hline
Map Synthesis                & \textit{Ganguli et al. \cite{ganguli2019geogan}}   & Google Map       & 2019          & GeoGAN            & \begin{tabular}[c]{@{}l@{}}\textit{Takes zoom level and resolution as input in addition to the feature}\\\textit{image}\end{tabular}                                                                                                                                                                                                                           \\ 
\hline
Map Synthesis                & \textit{Zhang et al. \cite{zhang2020enhanced}}     & GPS              & 2020          & SG-GAN            & \begin{tabular}[c]{@{}l@{}}\textit{Takes GPS coordinate as input in addition to the feature image}\end{tabular}                                                                                                                                                                                                                                      \\ 
\hline
Map Synthesis                & \textit{Andrade et al. \cite{andrade2020synthesis}}   & Historic map     & 2020          & CGAN              & \begin{tabular}[c]{@{}l@{}}\textit{Tried to convert historical maps into satellite view images for}\\\textit{ the first time.}\end{tabular}                                                                                                                                                                                                                      \\ 
\hline
Map Synthesis                & \textit{Ingale et al. \cite{ingale2021image}}    & Google Map       & 2021          & CGAN              & \begin{tabular}[c]{@{}l@{}}\textit{Used a set of unique GAN configurations tricks including}\\\textit{~normalizing the image between -1 to 1, using tanh as the last}\\\textit{layer of generator output, batch normalization for GAN training}\\\textit{~and avoiding sparse gradients such as RELU and maxpool}
\\ \textit{to achieve GAN stability}\end{tabular}  \\ 
\hline
Map Synthesis                & \textit{Song et al. \cite{song2021mapgen}}      & Google Map       & 2021          & MapGen-GAN        & \begin{tabular}[c]{@{}l@{}}\textit{Specialized in the conversion of remote-sensing images}\\\textit{to maps to be used in emergency response scenarios}\end{tabular}                                                                                                                                                                                            \\ 
\hline
Object Augmentation & \textit{Huang et al. \cite{huang2021object}}     & HRSC2016         & 2021          & CGAN              & \textit{Augmentation of weak objects like ships and boats in harbor scenes}                                                                                                                                                                                                                                                                                      \\ 
\hline
Object Augmentation & \textit{Martinson et al. \cite{martinson2021training}} & xView            & 2021          & CycleGAN          & \textit{Augmentation of 3D objects into satellite view images}                                                                                                                                                                                                                                                                                   \\
\hline
\end{tabular}
\end{table*}

The rest of the paper is organized as follows. Section II reviews related work. Section III provides background on related research. Section IV presents the proposed method. Section V demonstrates experimental setups and results. Section VI summarizes the results with related discussions, and finally, Section VII concludes the paper. \\

\section{Related Work}
Previous work on GAN-based satellite image synthesis can be categorized into two different types: (i) map synthesis; and (ii) object augmentation, as shown in table \ref{map}
\subsection{Map Synthesis}
Map generation is very important in emergency rescue operations such as responses to tsunamis, landslides, and earthquakes in which normal infrastructures like roads and buildings have been affected. In such conditions, the translation of satellite images to the map is a game changer for remediation planning and response. The majority of CGAN-based satellite image synthesis efforts concentrate on map synthesis.

\subsection{Object Augmentation}
Most of the satellite image datasets suffer a naturally imbalanced distribution of objects in the scene. Most of the pixels can be classified as roads, buildings, green areas, and water resources. For this reason, target objects like airplanes or ships are called weak objects. Object augmentation is the practice of inserting more weak objects into an already-available satellite image to address the imbalanced object distribution in satellite images.
 Martinson et al. \cite{martinson2021training} tried to insert 3D objects into satellite images using Blender3D given a specified viewing angle, lighting condition, and shadow. Huang et al. \cite{huang2021object} proposed an object-level remote sensing image augmentation to augment ships and boats to harbor scenes using CycleGAN.

\begin{figure*}[]
\centering
  \includegraphics[width=100mm]{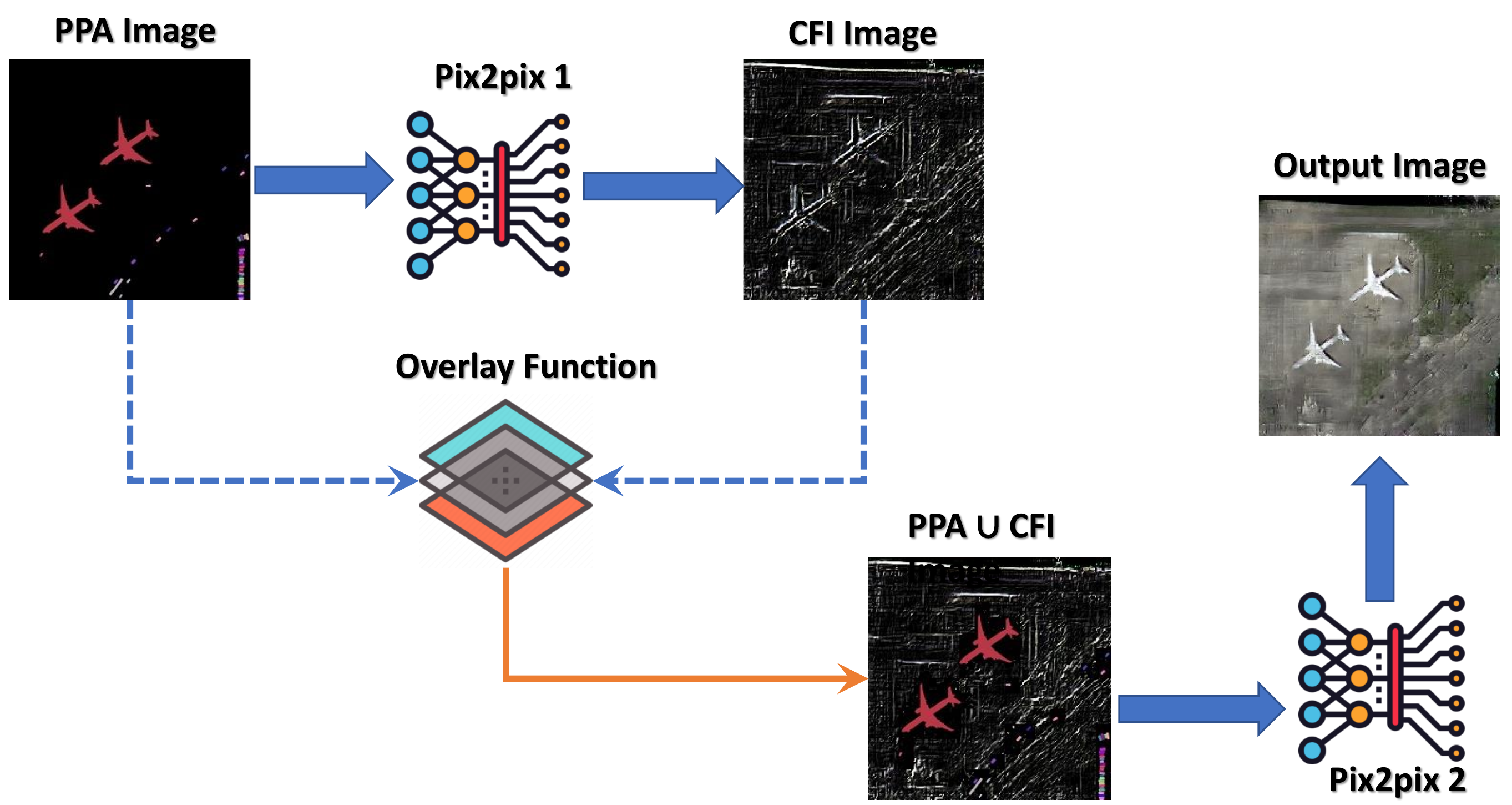}
  \caption{PDA image synthesis pipeline.}
  \label{pda}
\end{figure*}
\section{Background}

\subsection{Vanilla GAN}
The initial version of GANs \cite{goodfellow2014generative} is known as Vanilla GAN. The learning process of the Vanilla GANs is to train a discriminator $D$ and a generator $G$ simultaneously. The target of $G$ is to learn the distribution $p_g$ over data $x$. $G$ starts from sampling input variables $z$ from a uniform or Gaussian distribution $p_z(z)$, then maps the input variables $z$ to data space $G(z; \theta_g)$ through a differentiable network where $\theta_g$ represents network parameters. $D$ is a classifier $D(x; \theta_d)$ that aims to recognize whether the input is from training data or from $G$ where $\theta_d$ represents network parameters. The minimax objective for GANs can be formulated as follows:
\begin{dmath}
_{G}^{min} \quad_{D}^{max} \quad V_{GAN} (D,G) = \mathbb{E} _ {x \sim p_{x} } [log  D(x)]  + \quad \mathbb{E} _ {z \sim q_{z} } [log  (1- D(G(z)))] 
\end{dmath}
\subsection{Conditional GAN}
With Conditional GANs (CGANs) \cite{mirza2014conditional}, labels act as an extension to the latent space $z$ to generate and discriminate images better.
The objective function of CGANs is as follows:
\begin{dmath}
\min_G\max_DV(D,G)=\mathbb{E}_{x\sim p(data)(x)}[\log D_({_X|_Y} _)]+
\mathbb{E}_{Z\sim pz}(z))[\log (1-D(G(_Z|_Y)))]
\end{dmath}

\subsection{Pix2Pix}
A Pix2pix \cite{isola2017image} image is a conditional GAN that uses feature images as labels to be translated to target images. Pix2pix uses U-Net \cite{zhang2018road,kohl2018probabilistic} as the architecture of generator PatchGAN for its discriminator architecture. The most important feature of the Pix2pix generator is skip connections between each layer $i$ and layer $n - i$ to concatenate all channels at layer $i$ with those at layer $n - i$, where $n$ is the total number of layers.

\section{Proposed Method}
This section presents the PPA problem definition and two methods to tackle this problem. The core idea behind both proposed methods is to augment detail to the feature images first, then translate the new feature image to the output image. To do so, we use two different detail augmentation approaches: (i) Partial Detail Augmentation (PDA) and (ii) Full Detail Augmentation (FDA). In the next sub-sections, we demonstrate each one separately.

\subsection{Problem Definition}
The capacity of the generator in a CGAN is defined as its ability to map the input data from lower dimensionality to higher dimensionality. In other words, the larger distribution is mapped onto the simpler, lower dimensional distribution, and the generator network is used as an approximation for the remapping process. Consequently, the precision of the generator's capture of the manifold is directly related to the level of detail in the input image. This means that the less detail in the input image, the less detail in the output image that can be produced. The low-detail to low-detail image translation has been practiced before, as shown in Figure \ref{fig2}. This is a low-detail to low-detail image translation since the background pixels are not taken into account. However, in a low-detail to high-detail image translation, has never been investigated before, the background detail is approximated as well. We therefore take the first step in this direction, as demonstrated in the following sections.  

\begin{figure*}[]
\centering
  \includegraphics[width=100mm]{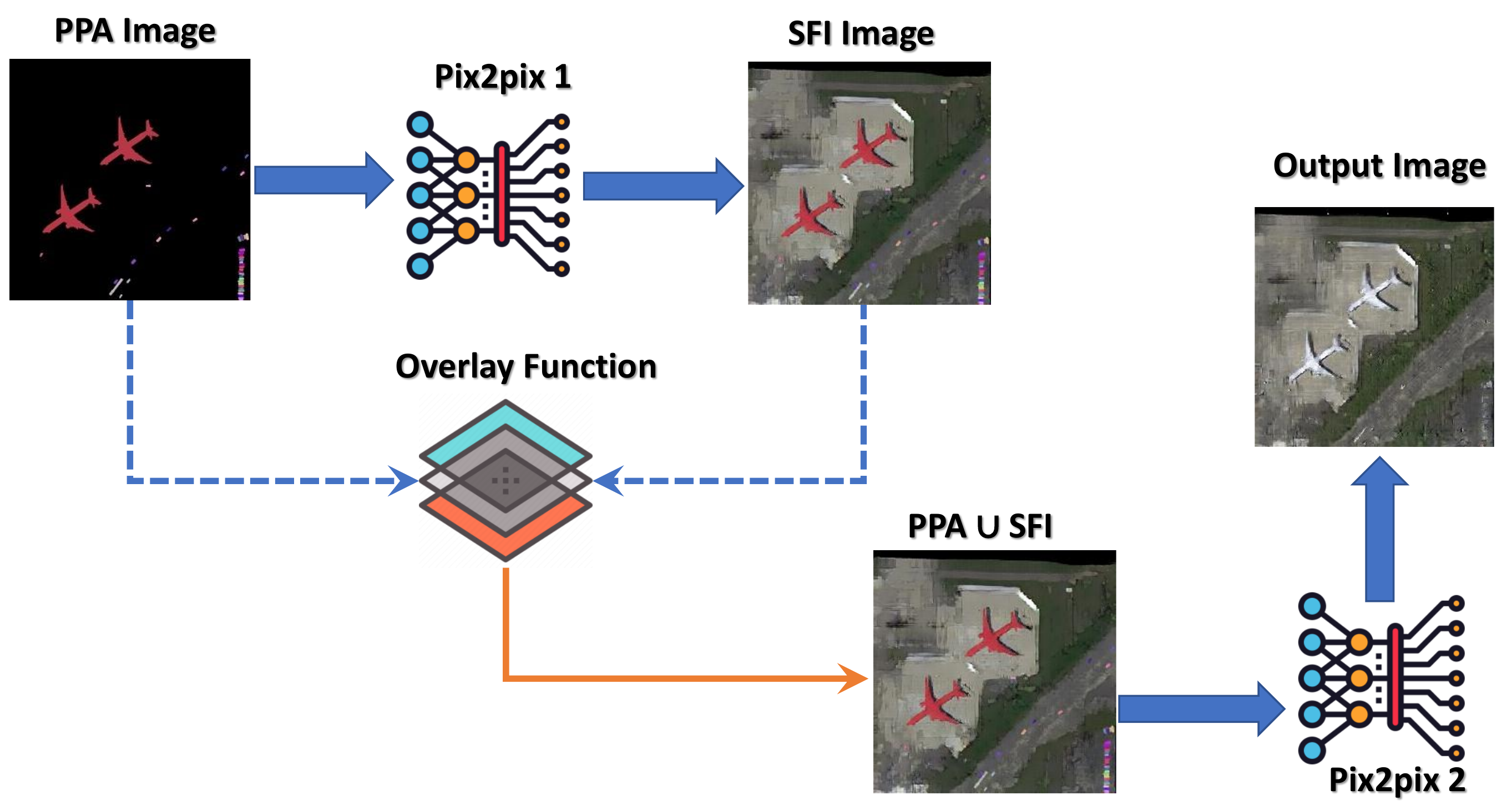}
  \caption{FDA image synthesis pipeline.}
  \label{fda}
\end{figure*}

\begin{figure}[H]
\centering
  \includegraphics[width=55mm]{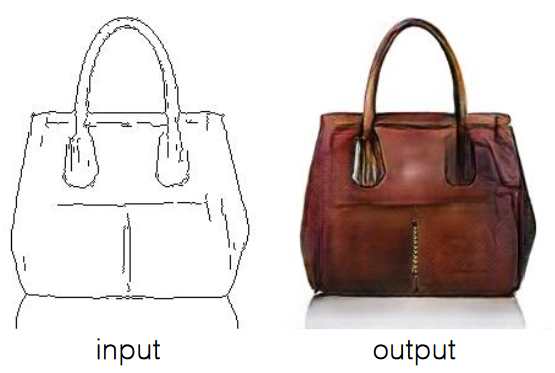}
  \caption{Low detail to low detail image translation using pix2pix \cite{isola2017image}}
  \label{fig2}
\end{figure}
\subsection{Partial Detail Augmentation (PDA)}

In this approach, we augment the minimum possible details of the feature images. Arguably the Canny edges fit this description, and so we add the Canny edges to the regions with missing color annotation. First, we train a Pix2pix model to translate the input feature images to a Canny Feature Image (CFI). Afterward, we overlay the initial color pixel-wise annotations on obtained CFI. Finally, we train a second Pix2pix model to translate the new feature image containing both color and Canny edges to the final output, as shown in Figure \ref{pda}. Algorithm 1 shows the steps involved.

\begin{algorithm}[H]
\caption{ Partial Detail Augmentation (PDA) }
\begin{algorithmic}[1]
       \STATE{Pre-process the training data via extracting CFI for each training data.}
      \STATE{Train Pix2pix(1) to translate each PPA image to its CFI version.} 
      \STATE{ Overlay PPA images on their corresponding CFI. } 
      \STATE{Train Pix2pix(2) to translate obtained images from Step 2 to output image.}

\end{algorithmic}
\end{algorithm}

\subsection{Full Detail Augmentation (FDA)}

In this approach, we augment the maximum possible details to the feature images. In fact, a specific color is assigned to each pixel to reach the maximum level of details in the background scene of the output image with missing color annotation. First, we train a Pix2pix model to translate the input feature images to a Segmented Feature Image (SFI). Afterward, we overlay the initial color pixel-wise annotations on obtained SFI. Finally, we train a second Pix2pix to translate the new feature image containing both color and Canny edges to the final output, as shown in Figure \ref{fda}. Algorithm 2 shows the steps involved.

\begin{algorithm}[H]
\caption{ Full Detail Augmentation (FDA) }
\begin{algorithmic}[1]
      \STATE{Pre-process the training data via extracting SFI for each training data and overlay the color pixel from PPI on corresponding SFI.}
      \STATE{Train Pix2pix(1) to translate each PPA image to its SFI version.} 
      \STATE{ Overlay PPA images on their corresponding SFI. } 
      \STATE{Train Pix2pix(2) to translate obtained images from Step 2 to output image.}

\end{algorithmic}
\end{algorithm}


\section{Experiments}
This section presents the experimental procedures, metrics, datasets, and related comparisons.

\subsection{Dataset}
In this paper, we take advantage of iSAID \cite{waqas2019isaid} as one of the rare pixel-wise annotated satellite image datasets to employ for tackling the PPA problem in CGAN-based image synthesis. According to \cite{waqas2019isaid}, "existing Earth Vision datasets are either suitable for semantic segmentation or object detection. ISAID is the first benchmark dataset specifically focused on the segmentation of aerial images. This large-scale and densely annotated dataset contains 655,451 object instances in 15 categories within 2,806 high-resolution images. The distinctive characteristics of iSAID are the following:" (a) a large number of images with high spatial resolution; (b) 15 important and commonly occurring categories; (c) a large number of instances of each category; (d) a large count of labeled instances per image, which might help in learning contextual information; (e) huge object scale variation, containing small, medium and large objects, often within the same image; (f) imbalanced and uneven distribution of objects with varying orientation within images, depicting real-life aerial conditions; (g) several small size objects, with ambiguous appearance, can only be resolved with contextual reasoning; and (h) precise instance-level annotations carried out by professional annotators, cross-checked and validated by expert annotators complying with well-defined guidelines.

\subsection{Object Detection Models }
We used three different target models as follows.
\begin{itemize}
\item AWS Rekognition: A cloud-based software as a service computer vision platform developed by Amazon.
\item  Cloud Vision API: A cloud-based software-as-a-service (SaaS) computer vision platform developed by Google.
\item YOLOv3: One of the most popular open-source object detectors that identify specific objects in videos, live feeds, or images.  
\end{itemize}

\begin{figure*}[]
\centering

  \includegraphics[width=175mm]{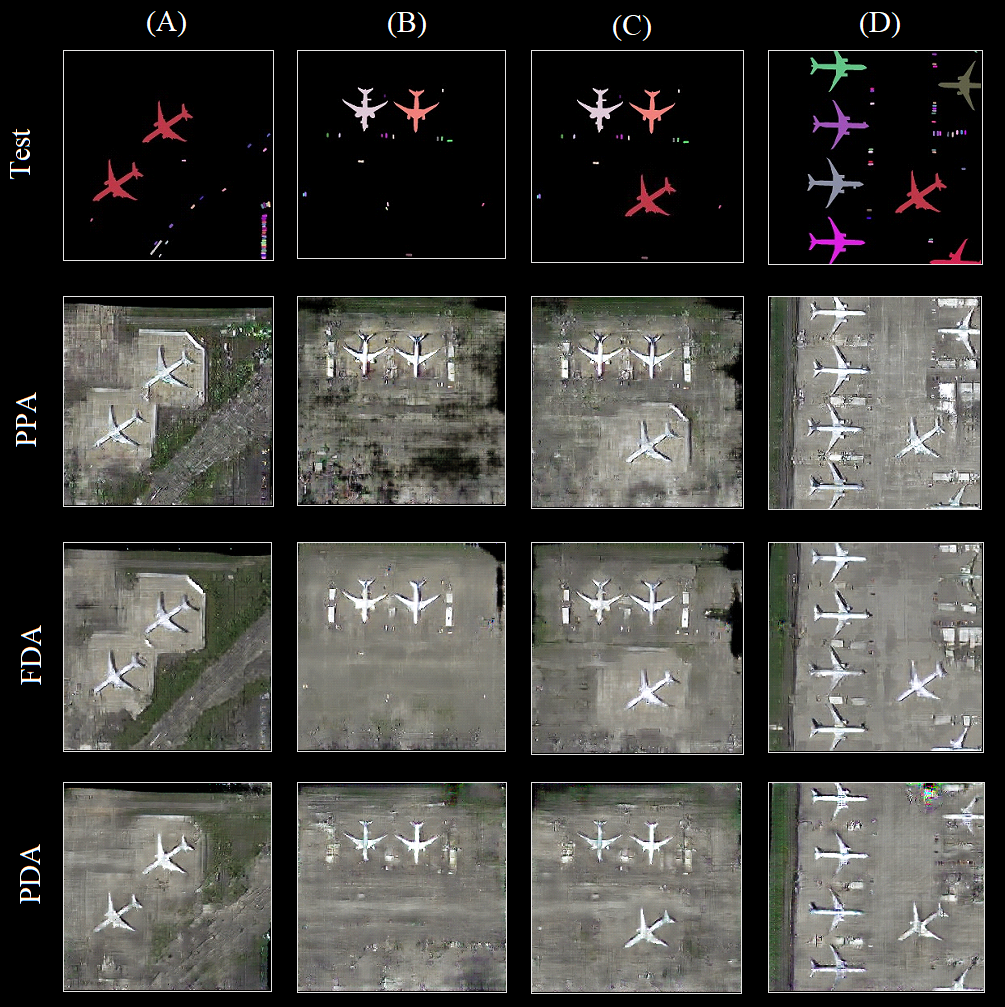}
  \caption{Unseen results generated by PPA (No detail augmentation), PDA, and FDA.}
  \label{unseen}
\end{figure*}

\begin{figure*}[]
\centering

  \includegraphics[width=175mm]{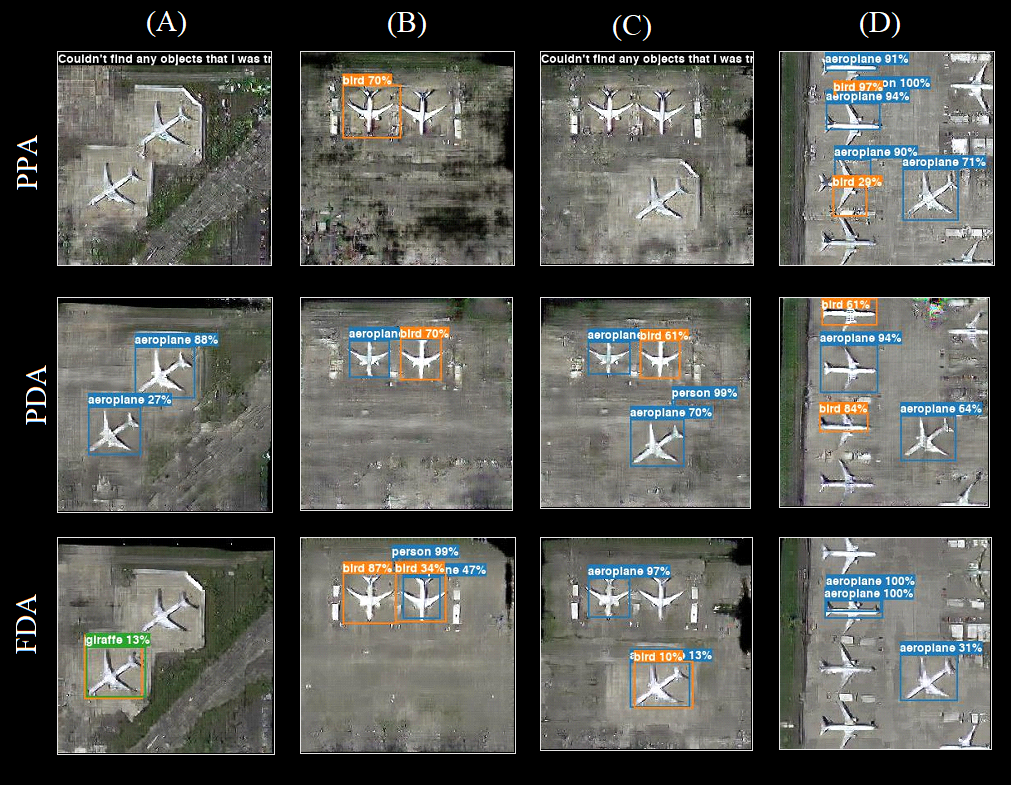}
  \caption{ Object Detection Scores using Yolov3 model: Three major labels are "Aeroplane", "Bird," and "Person".}
  \label{yolo}
\end{figure*}
\subsection{Metric}
To evaluate the quality of unseen results, we use the object detection score (ODS) returned from the three models mentioned earlier. This score shows (i) whether or not a target object has been detected, and (ii) the confidence of the model about the detected object. The higher ODS of target objects generally reflects higher quality of the synthesized object in the scene.

\subsection{Target Objects}
During our experiments, we focused on satellite-view airport scenes and related objects, including airplanes, aircraft, buildings, and vehicles. Since different infrastructures can be interpreted as buildings, we consider all related labels as "Building" like "Aerospace Manufacturer" detected mainly by the Google Vision tool. 
\subsection{Disclaimer}
All the ODS obtained from online tools like AWS Rekognition and Google Cloud were obtained on November 11, 2022. Considering the fact that these tools are constantly upgraded, different scores might be obtained in case of experiments in the future. However, using the dataset as posted on this date should allow replication of our experiments.
\section {Results}
This section summarizes the experimental results. During our experiments, we focused on synthesizing unseen results given input PPA test data as shown in Figure \ref{unseen}.
After generating the output, it is passed to three aforementioned object detectors. Tables II and III tabulate the returned ODS. Also, Figure \ref{yolo} shows the input test instances and corresponding YOLOv3 ODS results printed on the instances. The reported results can be summarized as follows.
\begin{table}[]
\centering
\caption{Object detection scores returned by Google Vision Tool.}
\begin{tabular}{c|cccc|cccc|}
\cline{2-9}
 & \multicolumn{4}{c|}{\textit{\textbf{Airplane}}} & \multicolumn{4}{c|}{\textit{\textbf{Aircraft}}} \\ \cline{2-9} 
 & \multicolumn{1}{c|}{\textbf{A}} & \multicolumn{1}{c|}{\textbf{B}} & \multicolumn{1}{c|}{\textbf{C}} & \textbf{D} & \multicolumn{1}{c|}{\textbf{A}} & \multicolumn{1}{c|}{\textbf{B}} & \multicolumn{1}{c|}{\textbf{C}} & \textbf{D} \\ \hline
\multicolumn{1}{|c|}{\textit{PPA}} & \multicolumn{1}{c|}{57.0} & \multicolumn{1}{c|}{0} & \multicolumn{1}{c|}{\textbf{76.0}} & \textbf{88.0} & \multicolumn{1}{c|}{70.0} & \multicolumn{1}{c|}{0} & \multicolumn{1}{c|}{58.0} & \textbf{91.0} \\ \hline
\multicolumn{1}{|c|}{\textit{PDA}} & \multicolumn{1}{c|}{54.0} & \multicolumn{1}{c|}{0} & \multicolumn{1}{c|}{0} & 78.0 & \multicolumn{1}{c|}{67.0} & \multicolumn{1}{c|}{0} & \multicolumn{1}{c|}{62.0} & 89.0 \\ \hline
\multicolumn{1}{|c|}{\textit{FDA}} & \multicolumn{1}{c|}{\textbf{90.0}} & \multicolumn{1}{c|}{0} & \multicolumn{1}{c|}{72.0} & 85.0 & \multicolumn{1}{c|}{\textbf{93.0}} & \multicolumn{1}{c|}{0} & \multicolumn{1}{c|}{\textbf{79.0}} & 90.0 \\ \hline
 & \multicolumn{4}{c|}{\textit{\textbf{Building}}} & \multicolumn{4}{c|}{\textit{\textbf{Vehicle}}} \\ \cline{2-9} 
 & \multicolumn{1}{c|}{\textbf{A}} & \multicolumn{1}{c|}{\textbf{B}} & \multicolumn{1}{c|}{\textbf{C}} & \textbf{D} & \multicolumn{1}{c|}{\textbf{A}} & \multicolumn{1}{c|}{\textbf{B}} & \multicolumn{1}{c|}{\textbf{C}} & \textbf{D} \\ \hline
\multicolumn{1}{|c|}{\textit{PPA}} & \multicolumn{1}{c|}{0} & \multicolumn{1}{c|}{0} & \multicolumn{1}{c|}{0} & \textbf{83.0} & \multicolumn{1}{c|}{0} & \multicolumn{1}{c|}{0} & \multicolumn{1}{c|}{0} & 91.0 \\ \hline
\multicolumn{1}{|c|}{\textit{PDA}} & \multicolumn{1}{c|}{0} & \multicolumn{1}{c|}{0} & \multicolumn{1}{c|}{57.0} & 51.0 & \multicolumn{1}{c|}{57.0} & \multicolumn{1}{c|}{0} & \multicolumn{1}{c|}{0} & 70.0 \\ \hline
\multicolumn{1}{|c|}{\textit{FDA}} & \multicolumn{1}{c|}{\textbf{79.0}} & \multicolumn{1}{c|}{0} & \multicolumn{1}{c|}{\textbf{58.0}} & 79.0 & \multicolumn{1}{c|}{\textbf{91.0}} & \multicolumn{1}{c|}{0} & \multicolumn{1}{c|}{0} & 91.0 \\ \hline
\end{tabular}
\end{table}

\begin{table}[]
\centering
\caption{Object detection scores returned by AWS Rekognition.}
\begin{tabular}{c|cccc|cccc|}
\cline{2-9}
 & \multicolumn{4}{c|}{\textit{\textbf{Airplane}}} & \multicolumn{4}{c|}{\textit{\textbf{Aircraft}}} \\ \cline{2-9} 
 & \multicolumn{1}{c|}{\textbf{A}} & \multicolumn{1}{c|}{\textbf{B}} & \multicolumn{1}{c|}{\textbf{C}} & \textbf{D} & \multicolumn{1}{c|}{\textbf{A}} & \multicolumn{1}{c|}{\textbf{B}} & \multicolumn{1}{c|}{\textbf{C}} & \textbf{D} \\ \hline
\multicolumn{1}{|c|}{\textit{PPA}} & \multicolumn{1}{c|}{84.1} & \multicolumn{1}{c|}{96.3} & \multicolumn{1}{c|}{88.5} & \textbf{95.2} & \multicolumn{1}{c|}{88.9} & \multicolumn{1}{c|}{96.3} & \multicolumn{1}{c|}{88.5} & \textbf{95.2} \\ \hline
\multicolumn{1}{|c|}{\textit{PDA}} & \multicolumn{1}{c|}{\textbf{89.3}} & \multicolumn{1}{c|}{\textbf{97.4}} & \multicolumn{1}{c|}{\textbf{91.4}} & 79.3 & \multicolumn{1}{c|}{\textbf{89.3}} & \multicolumn{1}{c|}{\textbf{97.4}} & \multicolumn{1}{c|}{\textbf{91.4}} & 79.3 \\ \hline
\multicolumn{1}{|c|}{\textit{FDA}} & \multicolumn{1}{c|}{57.2} & \multicolumn{1}{c|}{93.1} & \multicolumn{1}{c|}{91.1} & 93.7 & \multicolumn{1}{c|}{64.1} & \multicolumn{1}{c|}{93.1} & \multicolumn{1}{c|}{91.1} & 93.7 \\ \hline
 & \multicolumn{4}{c|}{\textit{\textbf{Building}}} & \multicolumn{4}{c|}{\textit{\textbf{Vehicle}}} \\ \cline{2-9} 
 & \multicolumn{1}{c|}{\textbf{A}} & \multicolumn{1}{c|}{\textbf{B}} & \multicolumn{1}{c|}{\textbf{C}} & \textbf{D} & \multicolumn{1}{c|}{\textbf{A}} & \multicolumn{1}{c|}{\textbf{B}} & \multicolumn{1}{c|}{\textbf{C}} & \textbf{D} \\ \hline
\multicolumn{1}{|c|}{\textit{PPA}} & \multicolumn{1}{c|}{0} & \multicolumn{1}{c|}{\textbf{57.0}} & \multicolumn{1}{c|}{0} & 0 & \multicolumn{1}{c|}{88.9} & \multicolumn{1}{c|}{96.3} & \multicolumn{1}{c|}{88.5} & \textbf{95.2} \\ \hline
\multicolumn{1}{|c|}{\textit{PDA}} & \multicolumn{1}{c|}{0} & \multicolumn{1}{c|}{0} & \multicolumn{1}{c|}{0} & 0 & \multicolumn{1}{c|}{\textbf{89.3}} & \multicolumn{1}{c|}{\textbf{97.4}} & \multicolumn{1}{c|}{\textbf{91.4}} & 79.3 \\ \hline
\multicolumn{1}{|c|}{\textit{FDA}} & \multicolumn{1}{c|}{0} & \multicolumn{1}{c|}{0} & \multicolumn{1}{c|}{0} & 0 & \multicolumn{1}{c|}{64.1} & \multicolumn{1}{c|}{93.1} & \multicolumn{1}{c|}{91.1} & 93.7 \\ \hline
\end{tabular}
\end{table}

\begin{figure*}[]
\centering

  \includegraphics[width=140mm]{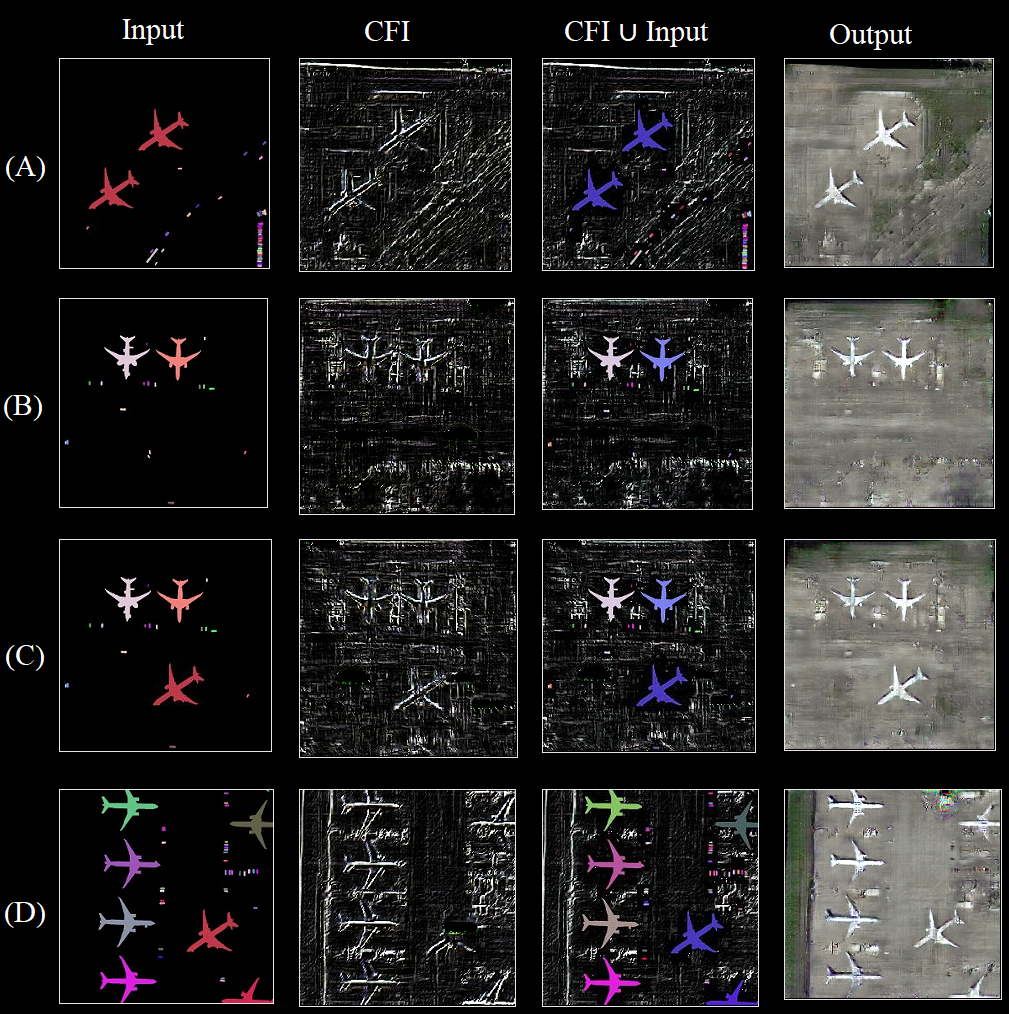}
  \caption{The outputs of PDA pipeline. The colors in ($Input \cup CFI$) were converted to increase the generalization power of second Pix2pix model.}
  \label{p5}
\end{figure*}

\begin{itemize}
\item PDA and FDA show their impact mostly on Building and Vehicle scores, according to ODS returned by Google Vision tool.
\item PDA and FDA show their impact on all scores except Building, according to ODS returned by AWS Rekognition.
\item PDA outperforms FDA in terms of all detected scores, according to ODS returned by AWS Rekognition.

\item PDA and FDA outperform PPA in terms of all detected scores, according to ODS returned by YOLOv3.
\item  PDA shows better performance in terms of true positive detected scores, according to ODS returned by YOLOv3.
\item FDA suffers a higher false positive rate compared to PDA and PPA, according to ODS returned by YOLOv3.
\end{itemize}
We also reported the outputs of the PDA pipeline in Figure \ref{p5}.
\section{Conclusion}
In this paper, we studied the partial pixel-wise problem in CGAN-based image synthesis. We discussed that a low-detail to high-detail image translation is not a trivial task since it directly impacts the capacity of CGAN models. To address this problem, we proposed two detail augmentation approaches called PDA and FDA. Our experimental results show that the proposed methods can enhance objection detection scores.
 \bibliographystyle{plain} 
\bibliography{conference_101719.bib}

\end{document}